\documentclass[10pt,journal]{IEEEtranTCOM}
\usepackage[english]{babel}
\usepackage[usenames]{color}
\usepackage[cp1250]{inputenc}
\usepackage{amsfonts}
\usepackage{amsthm}
\usepackage{graphicx}
\usepackage{epsfig}
\usepackage{mathrsfs}
\usepackage{amsmath}
\usepackage{algorithm}
\usepackage{algorithmic}
\usepackage{hyperref}

\theoremstyle{plain}

\begin{document}
\newcommand{\bea}{\begin{eqnarray}}
\newcommand{\eea}{\end{eqnarray}}
\newcommand{\be}{\begin{equation}}
\newcommand{\ee}{\end{equation}}
\newcommand{\beas}{\begin{eqnarray*}}
\newcommand{\eeas}{\end{eqnarray*}}
\newcommand{\bs}{\backslash}
\newcommand{\bc}{\begin{center}}
\newcommand{\ec}{\end{center}}
\def\SC {\mathscr{C}}

\title{Polynomial-based rotation invariant features}
\author{\IEEEauthorblockN{Jarek Duda}\\
\IEEEauthorblockA{Jagiellonian University,
Golebia 24, 31-007 Krakow, Poland,
Email: \emph{dudajar@gmail.com}}}
\maketitle

\begin{abstract}
One of basic difficulties of machine learning is handling unknown rotations of objects, for example in image recognition. A related problem is evaluation of similarity of shapes, for example of two chemical molecules, for which direct approach requires costly pairwise rotation alignment and comparison. Rotation invariants are useful tools for such purposes, allowing to extract features describing shape up to rotation, which can be used for example to search for similar rotated patterns, or fast evaluation of similarity of shapes e.g. for virtual screening, or machine learning including features directly describing shape. A standard approach are rotationally invariant cylindrical or spherical harmonics, which can be seen as based on polynomials on sphere, however, they provide very few invariants - only one per degree of polynomial. There will be discussed a general approach to construct arbitrarily large sets of rotation invariants of polynomials, for degree $D$ in $\mathbb{R}^n$ up to $O(n^D)$ independent invariants instead of $O(D)$ offered by standard approaches, possibly also a complete set: providing not only necessary, but also sufficient condition for differing only by rotation (and reflectional symmetry).
\end{abstract}
\textbf{Keywords: machine learning, feature extraction, computer vision, chemoinformatics, rotation invariants, spherical harmonics}
\section{Introduction}
Having a database of 2D or 3D objects, searching for them in real-life situations requires handling the difficulty of an unknown position, scale and rotation - for example in an image. While position and scale is relatively simple to normalize, e.g. by shifting to the center of mass and rescaling to a fixed average distance, unknown rotation is usually much more difficult to handle.

There are ways to normalize rotation of objects, e.g. by approximating with ellipsoid using PCA (principal component analysis) and rotating $k$-th longest axis to the $k$-th coordinate. However, an important issue of this approach is lack of continuity~\cite{shape}: often small modification can change e.g. order of ellipsoid radii, completely changing the description.

Hence, it is convenient to be able to extract features which do not change with rotation, allowing to directly compare objects in unknown rotations. For example in  virtual screening in chemoinformatics we know which ligands are activating given proteins - to use shape for such supervised learning, rotation-invariant features would allow to directly exploit shape as additional parameters deciding successfulness of a given molecule. Otherwise, pairwise comparing of shapes requires costly alignment and shape evaluation procedure for every pair of molecules - instead of inexpensive metric between vectors of rotation invariants.

A standard approach is using spherical harmonics to model spherical envelope: defining radius in every spherical angle. It uses complete basis, allowing to approximate spherical envelopes using series of coefficients - we can for example use such sequence of coefficients after PCA rotation normalization~\cite{PCAharm}. Alternatively, we can directly use rotation invariants: square averages of all coefficients for degree $d$ homogeneous polynomials~\cite{harm} on sphere, getting only one rotation invariant per degree $d$. \begin{figure}[b!]
    \centering
        \includegraphics{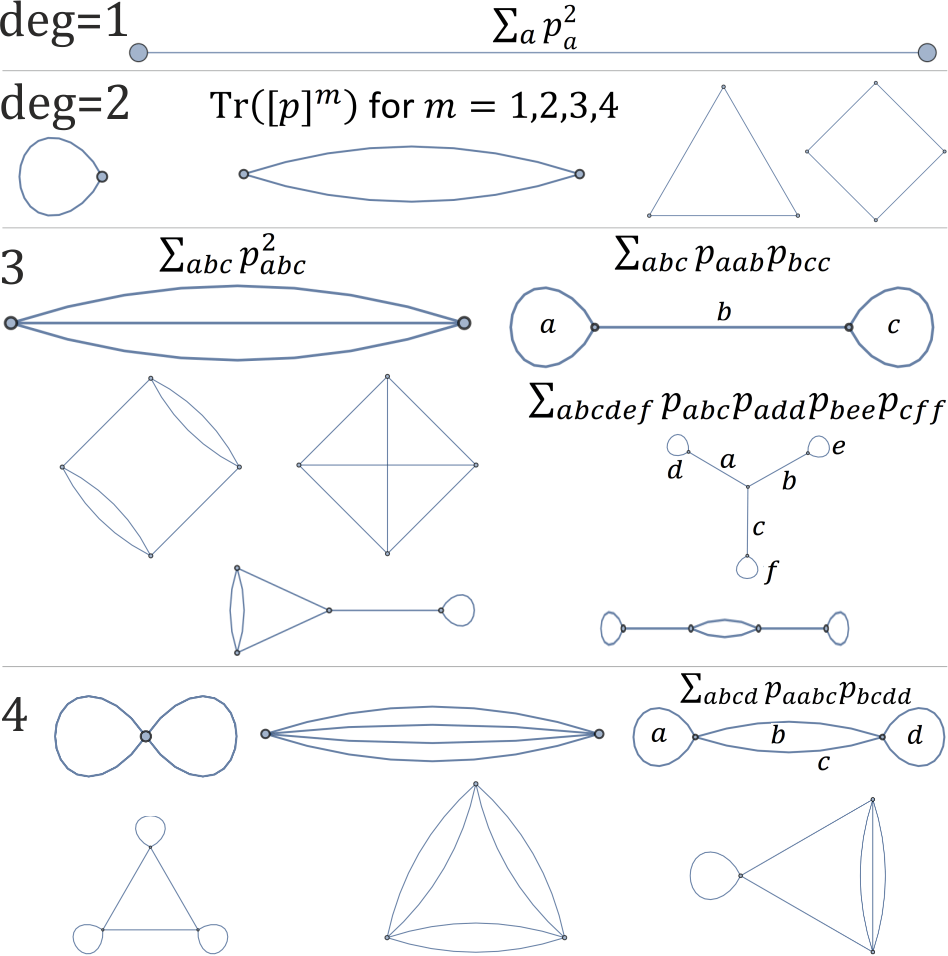}
        \caption{Diagrammatic representations of some first rotation invariants for degree 1, 2, 3, 4 homogeneous polynomials. Each vertex corresponds to a term of polynomial and has the same degree as polynomial. Operating on commutative fields like $\mathbb{R}$ here, edges for given vertex are indistinguishable. Every edge corresponds to summation over corresponding index, like in matrix product, and is rotation invariant thanks to $\sum_i O_{ai}O_{\alpha i}=\delta_{a\alpha}$ relation. Invariants from disconnected graphs can be omitted as being products over invariants for its components.}
       \label{diag}
\end{figure}

\begin{figure}[t!]
    \centering
        \includegraphics{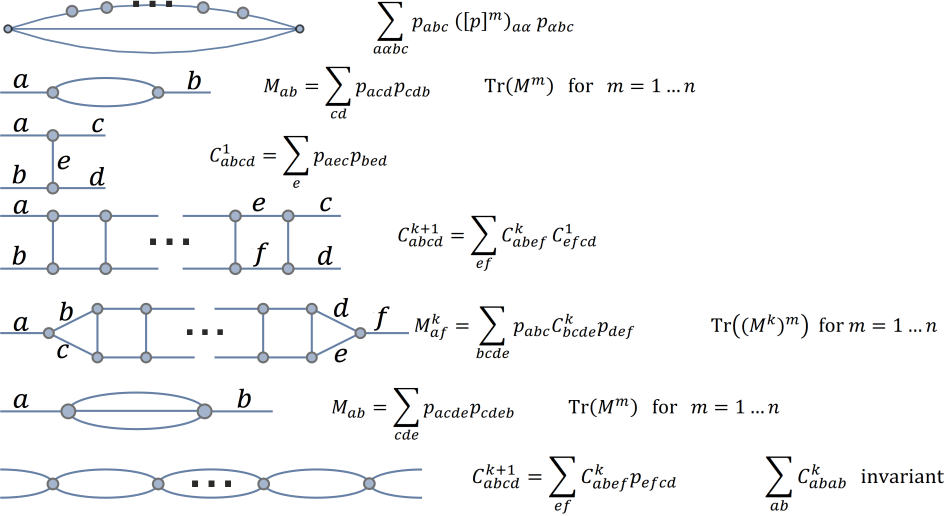}
        \caption{Some possibilities for systematic generation of large numbers of rotation invariants. For example through graphs with two external edges, which can be treated as matrices - getting $n$ invariants from $\textrm{tr}(M^m)$, corresponding to cyclic graphs built from $m$ copies of a given element. Presented ladder-like constructions allows to generate arbitrarily large number of invariants (also for higher degrees), however, there remains a difficult question of their independence. }
       \label{aut}
\end{figure}

In contrast, in $\mathbb{R}^n$ number of parameters suggests up to ${n+d-1 \choose d}-n(n-1)/2$ independent rotation invariants for degree $d$ homogeneous polynomial - there will be discussed their construction using diagrammatic representation like in Fig. \ref{diag}, \ref{aut}: each such graph corresponds to rotation invariant. However, efficient construction of complete set: determining polynomial up to rotation, remain an open question.

Beside additional invariants (comparing to cylindrical and spherical harmonics), and generalizing to arbitrary dimension, presented approach allow to work not only on sphere e.g. for spherical envelope describing relatively simple shapes, but also using general polynomials - directly describing e.g. 2D/3D density map, what seems more appropriate for many applications like chemical molecules or pixel maps.

\section{Rotation invariants for polynomials}
We will focus here on real degree $D$ polynomial $p:\mathbb{R}^n \to \mathbb{R}$, generally denoted by:
\be p(x) =\sum_{d=0}^D p^d(x)= p_\emptyset + \sum_i p_i x_i + \sum_{ij} p_{ij} x_i x_j +\ldots \label{pol}\ee
\noindent where $p^d(x)$ denotes homogeneous degree $d$ polynomial: $p^d(x)=\|x\|^d p^d(\hat{x})$, where $\|x\|=\sqrt{x^T x}$, $\hat{x}=x/\|x\|$. For example $p^1(x)=\sum_i p_{i\in I} x_i$ where $I=\{1,\ldots,n\}$ is default index range. Additionally, denote $[p]\equiv [p_{ij}]$ as the matrix for $p^2(x)$. The discussed approach is also valid for series ($D = \infty$).

Denote $x\to Ox$ as rotation for orthogonal $O^T O=O O^T=\mathbb{I}$. It modifies $x_i \to \sum_{ai} O_{ai} x_i$, what is equivalent to modification of polynomial coefficients:
\be p_{i}\to\sum_{a} p_{a} O_{ai} \qquad p_{ij}\to\sum_{ab} p_{ab} O_{ai} O_{bj} \label{trans}\ee
and so on - we are interested in constructing rotation invariants which remain fixed under such modification for all coefficients (using the same $O$):
\be p\sim q\qquad\equiv\qquad \exists_{O:O^TO=\mathbb{I}}\quad p(x)=q(Ox),\ee
what also includes reflective symmetries as $O$ can have $-1$ eigenvalues. Hence the presented approach is not sufficient to distinguish mirror versions of polynomials and so objects they represent, like left and right hand, or enantiomers of chiral molecules. However, these are just two possibilities, which can be inexpensively tested in some further stage.
\subsection{Homogeneous polynomials}
There are well known rotation invariants for degree 0, 1 and 2 homogeneous polynomials:
\begin{itemize}
\item If $p\sim q$, their 0-th terms have to agree: $p_\emptyset=q_\emptyset$.
  \item $p^1(x)=\sum_i p_i x_i$ degree 1 homogeneous polynomial, has single invariant for rotation $p_i\to \sum_a p_a O_{ai}$: $\sum_i p_i^2$, which completely characterizes it up to rotation: $p^1\sim q^1 \Leftrightarrow \sum_i p_i^2=\sum_i q_i^2$.
      $$\sum_i p_i^2=\sum_{ia\alpha} p_a O_{ai}\, p_\alpha O_{\alpha i}=\sum_{a\alpha}p_a p_\alpha \delta_{a\alpha}=\sum_a p_a^2 $$
  \item $p^2(x)=\sum_{ij}p_{ij} x_i x_j$ degree 2 homogeneous polynomial is just scaling in eigendirections of $[p]$ matrix of coefficients. $p^2\sim q^2$ iff their $n$ eigenvalues agree (with multiplicities), or equivalently $\{\lambda^0,\ldots,\lambda^{n-1}\}$ coefficients of characteristic polynomials agree: $\det([p]-\lambda \mathbb{I})=\det([q]-\lambda \mathbb{I})$, or equivalently $\textrm{Tr}\left([p]^m\right)=\textrm{Tr}\left([q]^m\right)\ (=\sum_i \lambda_i^m)$ for $m=1,\ldots,n$ .
\end{itemize}

Degree 3 homogeneous polynomial is analogously transformed: 
$$p(x)=\sum_{ijk} p_{ijk} x_i x_j x_k \qquad p_{ijk}\to\sum_{abc} p_{abc} O_{ai} O_{bj} O_{ck}$$
We can easily check that for example $\sum_{ijk}p_{ijk}p_{jki}=$
$$=\sum_{ijk}\left(\sum_{abc}p_{abc}O_{ai}O_{bj}O_{ck}\right)
\left(\sum_{\alpha\beta\gamma} p_{\beta\gamma\alpha} O_{\beta j}O_{\gamma k} O_{\alpha i}\right)= $$
\noindent $=\sum_{abc} p_{abc}p_{bca}$ is rotation invariant using the $\sum_i O_{ai} O_{\alpha i}=\delta_{a\alpha}$ relation for $ijk$ indexes.

Analogously we can construct such invariants by using exactly two copies of indexes we sum over, allowing for their diagrammatic representation - some examples are presented in Fig. \ref{diag}, some approaches for systematic way for construction of large numbers of such invariants are presented in Fig. \ref{aut}.

However, some of them might be dependent, e.g. $\sum_{ab} p_{aa}p_{bb}=\left(\sum_a p_{aa}\right)\left(\sum_b p_{bb}\right)$, what would be represented by disconnected graph - hence it is sufficient to focus on connected graphs in diagrammatic representation.

There are also more sophisticated dependencies, e.g. $\textrm{Tr}([p]^{n+1})$ can be calculated from $\textrm{Tr}([p]^{m})=\sum_i \lambda_i^m$ for $m=1,\ldots,n$. This is caused by the fact that $n$ equations often determine $n$ variables. However, it requires some independence, which is generally a complicated question.

And so for degree 3 and higher, while agreement of proposed invariants is necessary for $p^d\sim q^d$, getting a sufficient condition: a complete set of rotation invariants, seems a really difficult question. The number of independent parameters we can optimize with $O(n)$ matrix, like in $[p]=O^T \textrm{diag}(\lambda_i) O$ case, is $n(n-1)/2$. The number of parameters in symmetric matrix is $n(n+1)/2$, hence optimization over orthogonal matrices allows to reduce the number of independent parameters to their difference: $n$, what agrees with the number of eigenvectors. Degree $d$ homogeneous polynomial analogously have ${{n+d-1}\choose d}$ parameters, suggesting
\be {{n+d-1}\choose d}-\frac{n(n-1)}{2}\ee
maximal number of independent rotation invariants. However, important problem of finding such complete bases seems difficult. Systematic approaches like in Fig. \ref{aut} might bring a solution here, and like for degree $2$ there are probably various ways to effectively realize such complete basis.
\subsection{Symmetry of indexes}
Denote $\imath\in I^d$ as sequence of indexes of $p_\imath x_{\imath_1}\cdot\ldots\cdot x_{\imath_d}$ degree $d$ term. Operating on commutative field $(x_i x_j=x_j x_i)$ like $\mathbb{R}$, coefficients of permutated indexes have identical meaning. Denote $\ell=\mathcal{L}(\imath)\in \mathbb{N}^n$ as function enumerating appearances of indexes - such that:
\be x_\imath:= x_{\imath_1}\cdot\ldots\cdot x_{\imath_d}=x_1^{\ell_1}\cdot \ldots \cdot x_n^{\ell_n}=:x^\ell.\ee
A given $\ell$ sequence of powers corresponds to
\be N_\ell:={|\ell| \choose \ell_1,\ldots,\ell_n}=\frac{|\ell|!}{\ell_1!\cdot \ldots \cdot \ell_n!} \qquad\textrm{indexes } \imath,\ee
where $|\ell|=\sum_i \ell_i=d$. Let us emphasize sorted indexes:
\be I^d_\leq:=\{\imath\in I^d: \imath_1\leq \imath_2\leq \ldots \leq \imath_d\} \ee
and $\mathcal{I}(\ell)=\imath\in I^d_\leq$, such that $\mathcal{I}\circ \mathcal{L}$ is identity on $I^d_\leq$, $\mathcal{L}\circ \mathcal{I}$ is identity on $\mathbb{N}^n$.\\

It might seem we have a freedom for distributing coefficients between all $N_\ell$ permutations $\mathcal{L}^{-1}(\ell)$, what might lead to additional invariants. However, there is needed a rotation-invariant control of this distribution, in analogy to matrix symmetrization $([p]+[p]^T)/2$ for degree 2, which is maintained if using equal coefficients for all $N_\ell$ indexes:
\be p_\imath = \frac{P_{\mathcal{L}(\imath)}}{N_{\mathcal{L}(\imath)}} \ee
and operate on unique $P_\ell:=\sum_{\imath:\mathcal{L}(\imath)=\ell} p_\imath$ coefficients - depending on sequence of powers, allowing to write our polynomial in 3 equivalent ways:
\be p(x)= \sum_{\imath\in I^*} p_\imath x_\imath=
\sum_{\imath\in I^*_\leq} N_{\mathcal{L}(\imath)} p_\imath x_\imath=
\sum_{\ell\in \mathbb{N}^n} P_\ell x^\ell \label{pol1} \ee
where $I^*:=\bigcup_d I^d$, $I_\leq^*:=\bigcup_d I^d_\leq$.
\subsection{General polynomials}
Observe that analogously we can construct invariants for general $p=\sum_{d\leq D} p^d$ polynomials: using graphs like in fig. \ref{diag}, but with vertices of varying degrees, equal to degree of the corresponding term. The simplest invariant obtained this way is $\sum_{ab} p_a p_{ab} p_b$ and analogously $\sum_{ab} p_a ([p]^k)_{ab} p_b$. Generally we can insert such degree 2 vertex (or a few) inside any edge of a graph, like presented in top of fig. \ref{aut}.

These mixed terms (with varying degrees) intuitively describe relative angle between homogeneous parts of a given polynomial, what is missing e.g. in standard rotation invariants based on cylindrical or spherical harmonics.

For simplicity assume that 2nd degree $[p]$ of our polynomial has nondegenerated eigenspectrum $\lambda_1<\ldots<\lambda_n$, what allows to rotationally normalize $p$ in unique way:
\be p(x) = p_\emptyset + \sum_i p_i x_i + \sum_{i} \lambda_i x^2_i +\ldots \label{poln}\ee

After taking homogeneous invariants: $p_\emptyset$, $\sum_a p_a^2$, $\textrm{Tr}([p]^m)$ for $m=1,\ldots,n$, we see that there are still missing $n-1$ parameters of $p^1$: defining relative angle between 2-nd degree ellipsoid and 1-st degree shift $(\hat{p^1})$.

In this normalized form (\ref{poln}), mixed invariants:
$$\sum_{ab} p_a ([p]^m)_{ab} p_b=\sum_a \lambda^m_a p^2_a $$
for $m=0$ gives previous $\sum_a p_a^2$. The assumption of all $\lambda_a$ being different, makes that these invariants for all $m=0,\ldots,n-1$ uniquely determine all $p_a^2$ (as Vandermonde determinant is nonzero). It leaves freedom of sign of $p_a\in \mathbb{R}$, but 2-nd degree term is symmetric under $x_a \to -x_a$. 

Finally, for degree $D=2$ with nondegerated eigenspectrum of $[p]$, we see that $2n+1$ rotation invariants determine: $ p(x) =p_\emptyset + \sum_i \pm p_i x_i + \sum_{i} \lambda_i x^2_i$ normalized polynomial. Half of them (having the same parity of number signs) can be rotated one into another, leaving two possibilities differing by reflectional symmetry.

For degenerated eigenspectrum of $[p]$, the number of degrees of freedoms is reduced, e.g. for $[p]\propto \mathbb{I}$ we have "a ball on a stick" situation, defined modulo rotation by only 3 parameters: $p_\emptyset$, distance $\sum_a p_a^2$, and radius e.g. $\textrm{Tr}([p])$. Hence $2n+1$ invariants for non-degenerated cases seem also sufficient for degenerated special situations - as they are described by a smaller number of invariants.

For higher degree polynomials situation becomes more complicated. Assuming commutative field, degree $d$ homogeneous polynomial has ${n+d-1 \choose d}$ coefficients. Rotation is generally $n(n-1)/2$ parameters. Hence by comparing dimensions: for $p^d$ we can expect at most ${n+d-1 \choose d}-n(n-1)/2$ independent rotation invariants, the remaining $n(n-1)/2$ parameters of relative rotation should come from mixed terms - describing angles comparing to e.g. lower degree terms. 

For example degree 2 terms allow to insert degree 2 vertices in various edges of graphs like in top fig. \ref{aut}: $\sum_{abc}p_{abc}^2\to \sum_{a\alpha bc} p_{abc}p_{a\alpha}p_{\alpha bc}$. However, practical construction of complete sets of invariants seems a difficult problem. For example second order term can turn out spherically symmetric: $[p]\propto\mathbb{I}$, not emphasizing any direction. Hence relative rotation should be described with mixed rotation invariants using terms of various degrees.

\subsection{Relative rotation of multiple polynomials}
Frobenius inner product (analog to scalar product for matrices), which induces Frobenius norm:
$$\langle A,B\rangle_F:=\textrm{Tr}(AB^T)=\sum_{ij} A_{ij} B_{ij}$$ $$\|A\|_F^2=\textrm{Tr}(AA^T)=\sum_{ij} A_{ij}^2$$
is invariant to rotation: $A\to O^TAO,\ B\to O^TBO$. Hence, it describes relative rotation between two degree 2 homogeneous polynomials defined by these matrices.

For example having two linear spaces $\mathcal{A}$ and $\mathcal{B}$ of symmetric matrices defining elipsoids as $\{x:x^T A x=1\}$, we can use Frobenius inner product to translate geometry between them~\cite{pnp}. Treating such matrix as vector:
$$V(A):=(A_{11},\ldots,A_{nn},\sqrt{2}A_{12},\ldots,\sqrt{2}A_{n-1,n})$$
we can use Frobenius inner product as standard scalar product: $\langle A,B\rangle_F=V(A)\cdot V(B)$. Now for example performing Gram-Schmidt orthonormalization using such scalar product, if $\mathcal{A}$ and $\mathcal{B}$ differ only by rotation, this basis would be orthonormal for both of them.\\

Such considerations about describing relative rotation for degree 2 homogeneous polynomials $p(x)$ and $q(x)$ can be also taken to other degrees (and numbers of polynomials) by using diagrams like in fig. \ref{diag} with vertices corresponding to different polynomials. For example the simplest graphs for degree 1,2,3 homogeneous polynomials are $\sum_a p_a q_a$ (scalar product), $\sum_{ab} p_{ab} q_{ab}$ (Frobenius inner product) and $\sum_{abc} p_{abc} q_{abc}$. They can treated as standard scalar product if converting them into vectors in the analogous way:
\be V(p):=\left(\sqrt{N_{\mathcal{L}(\imath)}}\ p_\imath\right)_{\imath\in I^*_\leq}=\left(\sqrt{N_\ell}^{\, -1}\ P_\ell\right)_{\ell\in \mathbb{N}^n}. \ee
Using polynomials as approximations of objects, we can get invariants for their relative rotation this way, probably up to $n(n-1)/2$ in analogy to rotation invariants for mixed degrees. However, again constructing a complete basis of invariants (providing sufficient condition) is a difficult problem.

\section{Spherical case}
In the previous section we have focused on rotation invariants for polynomials describing situation in $\mathbb{R}^n$, like density map. However, e.g. for spherical envelope we are interested only in distance in every direction: function defined only on unit sphere $S^n=\{x\in\mathbb{R}^n:\|x\|=1\}$.

We can analogously model such function with a polynomial $p(x)$, while in fact being interested only in its values for $x\in S$. For this polynomial we can find rotation invariants exactly like in the previous section.

The only difference for spherical case is the number of independent terms to consider for such general degree $D$ polynomial. It turns out that it allows to focus only on using e.g. last two degrees: $p(x)=p^{D-1}(x)+p^D(x)$. Lower degree terms are already present there as for our sphere degree 2 polynomial: $\sum_i x_i^2=1$.

\begin{figure}[t!]
    \centering
        \includegraphics{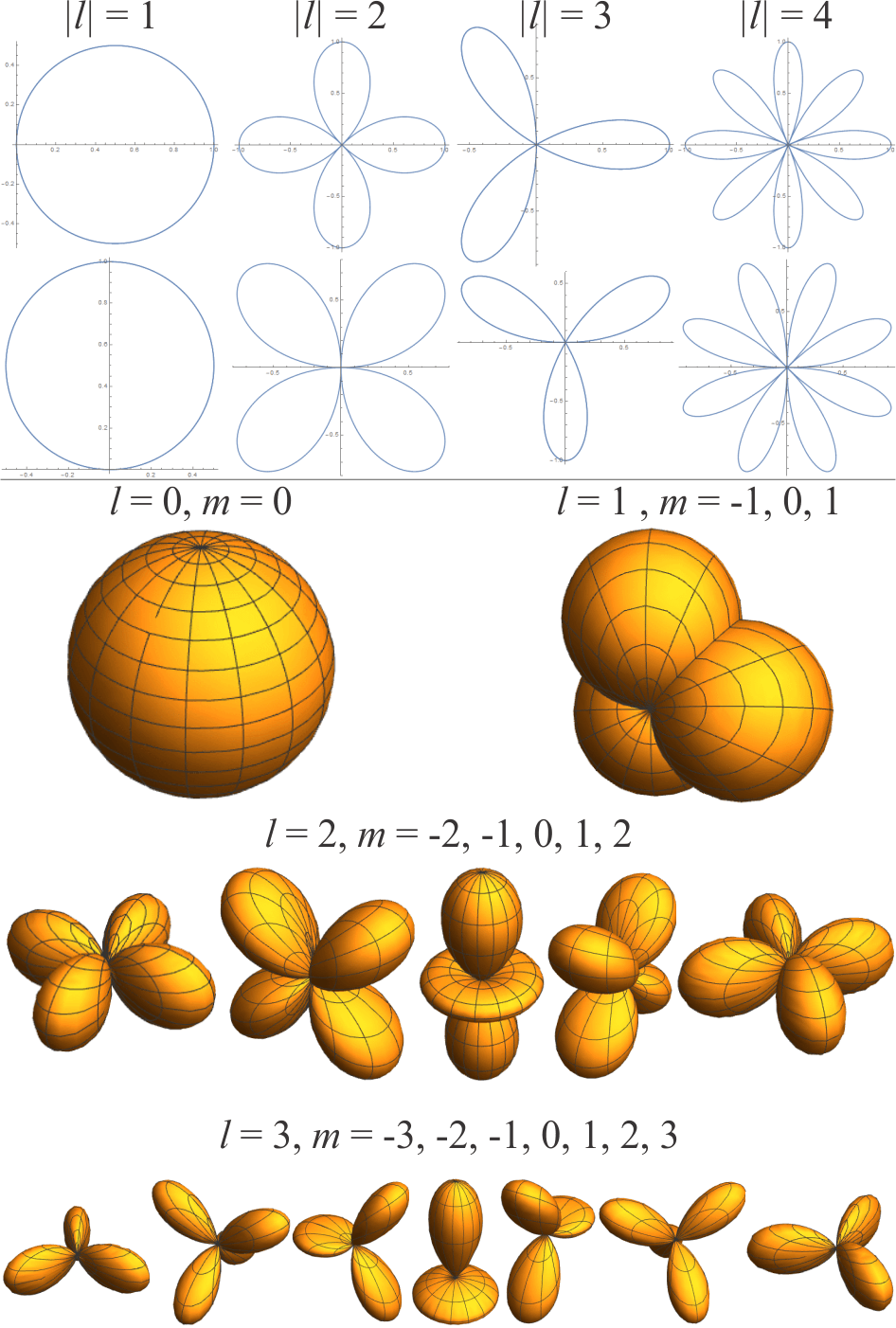}
        \caption{Top: cylindrical harmonics: $\cos(l\varphi)$, $\sin(l\varphi)$ in polar coordinates: $r=r(\varphi)$, for $|l|=1,2,3,4$. Bottom: real spherical harmonics in spherical coordinates: $r=r(\theta, \varphi)$, for $|l|=0,1,2,3$.}
       \label{harm}
\end{figure}

We will also relate to standard approaches: rotation invariants based on cylindrical and (real) spherical harmonics. They use orthonormal bases for $\langle f,g\rangle = \int_S fg dx$ scalar product, where $S$ is correspondingly $S^2$ or $S^3$. Some first of such functions are presented in fig. \ref{harm}.
\subsection{Cylindrical harmonics in 2D}

In $\mathbb{R}^2$ with $(x,y)$ coordinates, we naturally parameterize unit sphere with single angle $x=\cos(\varphi)$, $y=\sin(\varphi)$. Orthonormal basis for cylindrical harmonics is
$$f_0 = \frac{1}{\sqrt{2\pi}},\quad f_{+l}=\frac{1}{\sqrt{\pi}}\cos(l\varphi),\quad f_{-l}=\frac{1}{\sqrt{\pi}}\sin(l\varphi)$$
for $l=1,2,\ldots$. These sines and cosines can be expressed on $S^2$ as homogeneous degree $d=l$ polynomials of the original variables, getting first ($|l|=0,1,2,3$) being proportional correspondingly to:
$$1\ ;\ x,\ y\ ;\ x^2-y^2,\ 2xy\ ;\ x^3-3xy^2,\ 3x^2y-y^3.$$

They generate a complete $L^2$ basis on $S^2$, hence any function there, or equivalently of angle $\varphi$, can be approximated in this basis - as just Fourier series:
$$r(\varphi) = a_0 +\sum_{l=1}^D (a_{+l} \cos(l\varphi) + a_{-l} \sin(l\varphi))$$
for example defining spherical envelope described by distance to boundary of the region in every direction. It not necessarily has to be convex, e.g. "$\vee$" shape can be described this way, however, it can only describe relatively simple shapes: with single distance in every direction.

The number of coefficients is 2 per degree $d\geq 1$. In contrast, for $n=2$ the number of  terms of degree $d$ homogeneous polynomials is ${2+d-1 \choose d}=d+1$. We see that polynomials have $d-1$ more terms than cylindrical harmonics - this difference comes from defining behavior only on sphere. For example $x^2+y^2$ would be used by general polynomial, but it does not bring new dependance on $S^2$. Analogously $\left(\sum_i x_i^2\right) p(x)$ for all $p$ polynomials of degree $d-2$, were already determined by lower degree terms, finally getting $d+1 - (d-1)=2$ new terms per degree, exactly as for cylindrical harmonics.

Such Fourier expansion has 1 rotation invariant per degree: $A_l=a_{+l}^2+a_{-l}^2$, which uniquely describes contribution of $l$-th frequency, without providing its relative angle. Equality of $A_l$ invariants ensures differing only by rotation, hence the presented approach will not improve situation for homogeneous terms: there is just one rotation invariant per degree here. However, these invariants lack information about relative angles between such homogeneous polynomials  - one parameter per degree $l=d\geq 2$, getting $D-1$ additional rotation invariants, which might be constructed with discussed approach using mixed terms for the obtained polynomial of $(x,y)$ variables on $S^2$.

\subsection{Real spherical harmonics in 3D}
In $\mathbb{R}^3$ with $(x,y,z)$ coordinates, we can analogously use real spherical harmonics for unit sphere, or equivalently spherical angle $(\theta,\varphi)$: $x=\sin(\theta)\cos(\varphi)$, $y=\sin(\theta)\sin(\varphi)$, $z=\cos(\theta)$. They create complete basis, this time for degree $d=l$ there are $2l+1$ terms: for $m=-l,\ldots,+l$:

Real spherical harmonics form a complete orthonormal basis, useful e.g. for approximation of the distance in all directions from a fixed point of a spherical envelope:
$$ r(\theta,\varphi)=\sum_{l=0}^D \sum_{m=-l}^l a_{lm} f_{lm}(\theta,\varphi) $$

Analogously as for cylindrical harmonics, on unit sphere we can express them with homogeneous degree $d=l$ polynomials, ($l=0,1,2,3$) being proportional to\footnote{\url{https://en.wikipedia.org/wiki/Table_of_spherical_harmonics}}:
$$1\ ;\ y,\ z,\ x\ ;\ xy,\ yz,\ -x^2-y^2+2z^2,\ zx,\ x^2-y^2\ ;$$
$$(3x^2-y^2)y,\ xyz,\ 4(4z^2-x^2-y^2),\ z(2z^2-3x^2-3y^2),$$
$$x(4z^2-x^2-y^2),\ (x^2-y^2),\ (x^2-3y^2)x$$
Under rotation, $a_{lm}$ coefficients transform accordingly to the corresponding Wigner rotation matrices $R^l$:
$$a'_{lm} = \sum_{m'=-l}^l R^l_{mm'} a_{lm'}$$
which allows for $A_l=\sum_{m=-l}^l a_{lm}^2$ rotation invariants: only one per degree and without information about relative rotation between different degree terms. In contrast to these $D+1$ invariants, dimensionality suggests total of $D^2-2D-2$ rotation invariants, which can be constructed with the presented approach for the obtained polynomial of $(x,y,z)$ variables on $S^3$.

\subsection{Higher dimension spherical invariants}
The 3D spherical harmonics already require relatively complicated formulas, which seem difficult to generalize to higher dimensions. However, the $\langle f,g\rangle = \int_S fg dx$ orthonormality is not necessary to approximate a function with polynomial, and the discussed here invaraints work for any dimension.

Generally $p^d(x)$ may contain $\left(\sum_i x_i^2\right) q^{d-2}(x)$ terms, which on unit sphere are identical to $d-2$ degree $q^{d-2}$. It suggests to  use just the last two homogeneous terms for spherical cases as the lower ones are this way included there: 
$$r(\hat{x})=p(\hat{x})=p^{D-1}(\hat{x})+p^D(\hat{x}).$$

After fitting polynomial of this form to our data, we can use the discussed approach to construct rotation invariants: preferably ${n+D-1 \choose D}-n(n-1)/2$ for $p^D$, plus ${n+D-2 \choose D-1} -n(n-1)/2$ for $p^{D-1}$, plus $n(n-1)/2$ mixed invariants to describe their relative rotation. Finally, the number of independent rotation invariants should be ${n+D-1 \choose D}+{n+D-2 \choose D-1} -n(n-1)/2$, what is $O(n^D)$ instead of $O(D)$ offered by standard harmonics.

\section{Some examples of possible application}
Imagine we have prepared the set of terms: of all degrees up to $D$ for general case $(x^\ell)_{|\ell|\leq D}$ or for $D-1$ and $D$ degree for spherical case: $(x^\ell)_{|\ell|=D-1,D}$.

Before determining rotation invariants, it is crucial to normalize position and scale first, which have to at least approximately agree for both objects to test differing by rotation. A natural choice for position is shifting average position (some "center of mass") to zero. Regarding scale, in some situations distances are fixed, especially for chemical molecules there should be used the same scale for both compared objects. There are also cases where scaling is allowed, especially for image patterns, which scale depends on distance - in such situation it is crucial to normalize scale, e.g. to average distance being 1.

For mean square error (MSE) fitting, it is convenient (not necessary) to have prepared orthonormal basis for $\langle f,g\rangle =\int f(x)g(x) dx$, where integration is over the set of interest. In spherical case, for 2D, 3D situations we can use the known cylindrical or real spherical harmonics bases. However, integration over higher dimensional spheres is more difficult. In the general $\mathbb{R}^n$ case, integral of polynomial over the entire space is usually infinite. One way to handle it is limiting space to a finite ball. Another is (allowed) multiplying the polynomial by a radius dependent function, e.g. $f(x)=\exp(-\|x\|^2)\cdot p(x)$ or $f(x)=\exp(-\|x\|)\cdot p(x)$.

Having a fitted polynomial (eventually multiplied by a radius-dependent function), we can use the discussed methods to construct rotation invariants for polynomial $p$. Equality of invariants is a necessary condition for differing only by rotation. Getting sufficient condition might be also possible, but would require a large number $(O(n^D))$ of invariants, especially for high degree polynomials.

Some metrics for vectors of rotation invariants can be used to evaluate similarity of two shapes, e.g. of molecules for virtual screening in chemoinformatics. However, quantitative evaluation is difficult question and requires further work.

Another possibility is using these invariants as additional features describing shape of e.g. molecule, complementing information this way e.g. for supervised learning.
\subsection{Invariants for general polynomials}
Fitting a general polynomial is useful for representation of global objects, for example entire structure of chemical molecules, or visual 2D objects. Rotation invariants also allow to multiply polynomial by a function depending only on radius, like Gaussian $\exp(-\|x\|^2)\cdot p(x)$ or exponential $\exp(-\|x\|)\cdot p(x)$, which are convenient e.g. for modelling probability density and can be inexpensively estimated~\cite{rapid}.

After normalization, we have a set of pairs $(x^i,y^i)$, where $x^i$ is interesting position, $y^i$ is corresponding value we would like for our fitted function $(f(x^i)\approx y^i)$: assuming MSE optimization, minimize $\sum_i \|f(x^i)-y^i\|_2$. For image they are for example pairs of (position of pixel, its grayness). For molecules we can assume discrete points: pairs of position of atom and value as 1, or its atomic mass, or electron-negativity, or some other atomic parameter.

Now having orthonormal basis $(f_j)$, we can choose coefficient for $f(x)= \sum_i a_j f_j(x)$ as just sum $a_j =\sum_i f_j(x^i)$. However, such projection usually uses different scalar product than used for orthornormalization, e.g. discrete summation instead of integration. Hence, a safer approach is directly optimizing MSE without assuming orthogonality: for rectangular matrix $M=[f_j(x^i)]_{ij}$ and vector $b=(y^j)$, find vector $a$ minimizing $\|Ma-b\|_2$, what can be done using pseudo-inverse, or is directly implemented in popular numerical libraries. There is also suggested adding e.g. $c\sum_i a_i^2$ for minimization to reduce found coefficients.

Another possible application is testing if two sets of points differ only by rotations, what was the original motivation of the presented approach~\cite{pnp} for testing graph isomorphism through comparison of eigenspaces of adjacency matrix.
\subsection{Spherical invaraints}
Basic example for functions defined on sphere is spherical envelope: $\{x:\|x\|\leq r(\hat{x})\}$ where $\hat{x} =x/\|x\|$, which is a popular tool to describe a 3D shape. Another example is to describe texture e.g. on a ball, or other relatively simple set.

MSE fitting can be performed as previously, but using $\left(\hat{x^i}, \|x^i\|\right)$ set of points for fitting spherical envelope, or e.g. using greyness as value for fitting texture on a ball or other simple shape.

\section{Conclusions and further work}
Unknown rotation of objects is a basic problem of machine learning. This paper proposes a general methodology for constructing large family of rotation invariants, starting with fitting a polynomial (or e.g. $\exp(-\|x\|^2)\cdot p(x)$), this way enhancing possibilities offered by standard approaches like cylindrical and spherical harmonics to much larger number of rotation invariants, possibly up to a complete set: allowing to define polynomial up to rotation.

The main remaining question is efficient construction of complete sets of rotation invariants - sufficient to ensure that two polynomials differ only by rotation (and eventually reflectional symmetry). This question concerns mainly three situations: homogeneous polynomials, sum of two successive homogeneous polynomials for spherical case, and sum of all homogeneous polynomials up to a given degree $D$. This question can be split into understanding invariants for homogeneous polynomials, and of mixing terms ensuring fixed relative rotation between parts of different degrees.

Another difficult question regards using such rotation invariants to compare shapes of two objects like molecules - trying to evaluate similarity of two shapes as some distance between their vectors of invariants.

Finally, a complementing question is efficient search for rotation alignment for two polynomials differing only by rotation, or being close to it.

\bibliographystyle{IEEEtran}
\bibliography{cites}
\end{document}